\PassOptionsToPackage{table}{xcolor}
\documentclass[sigconf]{acmart}
\renewcommand\footnotetextcopyrightpermission[1]{} 
\setcopyright{none}

\usepackage[T1]{fontenc}
\usepackage{microtype}
\usepackage{graphicx}
\usepackage{subcaption}
\usepackage{booktabs}
\usepackage{colortbl}
\usepackage{multirow}
\usepackage{enumitem}
\usepackage[most]{tcolorbox}
\usepackage[capitalize,noabbrev]{cleveref}
\usepackage[textsize=tiny]{todonotes}
\usepackage{caption}
\tcbuselibrary{skins,breakable}

\definecolor{boxbody}{RGB}{255,255,255}
\definecolor{boxheader}{RGB}{0, 51, 102}
\definecolor{promptblue}{RGB}{70, 130, 180}
\definecolor{promptbg}{RGB}{248, 250, 252}
\definecolor{tagcolor}{RGB}{220, 20, 60}
\definecolor{lightgray}{gray}{0.9}

\theoremstyle{plain}

\theoremstyle{definition}

\theoremstyle{remark}

\definecolor{bg-gray}{gray}{0.92}
\definecolor{bg-green}{rgb}{0.96, 1.0, 0.98}

\AtBeginDocument{%
  }

\setcopyright{acmlicensed}
\copyrightyear{2018}
\acmYear{2018}
\acmDOI{XXXXXXX.XXXXXXX}
\acmISBN{978-1-4503-XXXX-X/2018/06}
\acmSubmissionID{}

\begin{document}

\title[EmoAgent-R1]{EmoAgent-R1: Towards Multimodal Emotion Understanding with Reinforcement Learning-based Dynamic Agent Specialization}
\author{Lihuang Fang}
\authornote{Both authors contributed equally to this research.}
\affiliation{%
  \institution{Guangdong University of Technology}
  \city{Guangzhou}
  \country{China}
}
\affiliation{%
  \institution{Southern University of Science and Technology}
  \city{Shenzhen}
  \country{China}
}
\email{l.h.fang228@gmail.com}

\author{Yuchen Zou}
\authornotemark[1]
\affiliation{%
  \institution{Xi'an Jiaotong University}
  \city{Xi'an}
  \country{China}
}
\email{yuchenzou@stu.xjtu.edu.cn}

\author{Kebing Jin}
\affiliation{%
  \institution{Guizhou Provincial Laboratory of Big Data, State Key Laboratory of Public Big Data, Guizhou University}
  \city{Guiyang}
  \country{China}
}
\email{kbjin@gzu.edu.cn}

\author{Jinghui Qin}
\authornote{Corresponding author.}
\affiliation{%
  \institution{Guangdong University of Technology}
  \city{Guangzhou}
  \country{China}
}
\email{qinjinghui@gdut.edu.cn}
\renewcommand{\shortauthors}{Fang et al.}

\begin{abstract}
Multimodal large language models (MLLMs) have achieved impressive performance in multimodal emotion recognition (MER) tasks and lifted MER to a new level that is complex emotion understanding with advanced video understanding abilities and natural language description.
However, existing MLLM-based methods often use a fixed prompt to perceive the emotions, ignoring the dynamicity and complexity of the emotion source in the multimodal inputs. 
To address these issues, we propose a novel Reinforcement Learning-based Dynamic Agent Specialization framework (\textbf{EmoAgent-R1}) to optimize the emotion recognition, reasoning, and generalization abilities of an MLLM with dynamic agent specialization based on reinforcement learning. Specifically, we first adopt a cold start strategy to endow an MLLM with preliminary emotion recognition, reasoning, and agent routing ability by training with synthetic answer-conditioned chain-of-thought data and agent routing data. Then, we further train the MLLM with reinforcement learning to perceive emotions in a two-step agentic workflow with agent selection and agent specialization. To effectively train EmoAgent-R1, we propose a novel Progressive Group-Relative Policy Optimization (P-GRPO) to combine group-based relative advantages with a PMI-inspired progressive token-level modulation to transform sparse rewards into fine-grained learning signals, mitigating the coarse-grained uniform credit assignment issue in GRPO. Extensive experiments on MER benchmarks demonstrate the superiority of our EmoAgent-R1 in stronger emotion reasoning performance and improved optimization stability.
\end{abstract}


\maketitle

\begin{figure*}[t]
    \centering
    \includegraphics[width=0.85\linewidth]{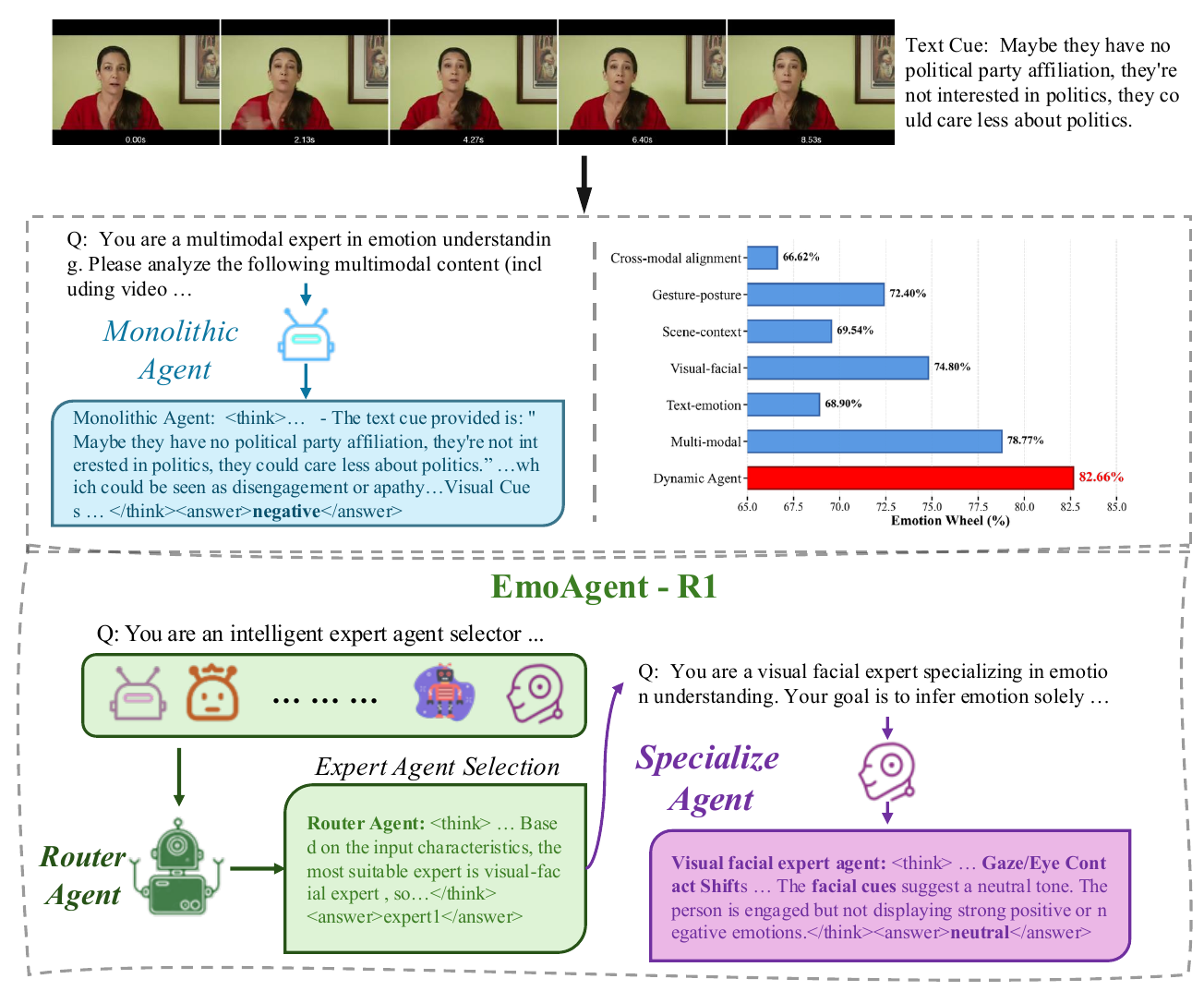}
    \caption{Comparison between a conventional monolithic agent (top part)  and the proposed EmoAgent-R1 framework (bottom part). The monolithic agent uses a single, static prompt to analyze all modalities uniformly. In contrast, EmoAgent-R1 employs a two-stage agentic workflow: a Router Agent first selects the most suitable specialized agent based on the input's characteristics, and then the selected Specialized Agent performs focused reasoning to derive the final emotion. This dynamic specialization enables more adaptive and accurate multimodal emotion understanding.}
    \label{fig:enter-label}
\end{figure*}

\section{Introduction}

Understanding human emotions from video is central to affective computing and pivotal for applications ranging from social media analysis to mental health assessment. Unlike unimodal sentiment analysis, Multimodal Emotion Recognition (MER) demands complex reasoning over heterogeneous and asynchronous signals, including facial expressions, vocal intonation, linguistic content, and temporal dynamics, etc. Although Multimodal Large Language Models (MLLMs) have significantly elevated MER performance by leveraging massive pretraining and advanced instruction tuning~\cite{cheng2024emotion, lin2024video,li2025emoverse}, they still suffer from a fundamental limitation: \textbf{they typically employ simplistic, static prompts with a monolithic pattern to perceive emotions, thereby neglecting the inherent dynamicity and complexity of emotional sources.}

In real-world scenarios, emotional cues are highly localized, sparse, and modality-dependent. For instance, a fleeting shift in vocal tone may dictate the sentiment even when visual expressions remain neutral, whereas sarcasm relies heavily on the contradiction between textual content and facial micro-expressions. Existing MLLM-based methods, which treat all modalities and temporal segments uniformly under a generic instruction, struggle to capture these shifting dynamics. This ``uniformity bias'' leads to brittle optimization and hallucinations, as the model lacks the flexibility to dynamically route its attention to the most informative evidence source. As illustrated in Fig.~\ref{fig:enter-label}, this limitation stems from the prevalent monolithic agent design in existing MLLM-based approaches. By relying on a single static prompt, such models enforce a uniform reasoning policy over all modalities and temporal segments, implicitly assuming that emotional cues are evenly distributed and equally informative. In contrast, real-world emotion understanding requires dynamically identifying which modality or temporal evidence is most salient for a given instance. This observation motivates an agentic workflow that decomposes emotion understanding into an explicit routing stage and a specialized reasoning stage, rather than a one-size-fits-all inference paradigm.

Reinforcement Learning (RL) offers a promising avenue to endow models with such dynamic reasoning capabilities. Recent advances, such as Group Relative Policy Optimization (GRPO)~\cite{shao2024deepseekmathpushinglimitsmathematical}, have demonstrated success in enhancing reasoning by optimizing policies directly against task-level rewards. However, applying standard RL to multimodal emotion understanding encounters two critical hurdles. First, \textbf{the monolithic policy limitation}: a monolithic policy is forced to master all reasoning patterns, failing to specialize in distinct emotional aspects (e.g., visual-dominant vs. text-dominant scenarios). Second, \textbf{the coarse-grained credit assignment issue}: reward signals are typically sparse and provided only at the sequence level. Standard GRPO broadcasts this scalar reward uniformly to all tokens, failing to distinguish critical reasoning steps from irrelevant noise, which results in inefficient exploration and unstable convergence in complex multimodal tasks.

To address these challenges, we propose \textbf{EmoAgent-R1}, a novel framework that achieves robust emotion understanding through \textbf{Reinforcement Learning-based Dynamic Agent Specialization}. We argue that effective emotion reasoning requires both \textit{structured specialization} and \textit{fine-grained credit assignment}. Specifically, our approach unfolds in two phases. First, recognizing the difficulty of learning complex reasoning from scratch, we adopt a \textbf{cold start training strategy} by constructing a dataset of synthetic answer-conditioned Chain-of-Thought (CoT) and agent routing data to endow the MLLM with preliminary capabilities in emotion reasoning and dynamic agent selection. Then, we further optimize the model using Reinforcement Learning with a novel \textbf{Progressive Group-Relative Policy Optimization (P-GRPO)}, which can overcome the limitations of standard GRPO. 
P-GRPO mitigates the coarse-grained uniform credit assignment issue in GRPO by combining group-based relative advantages with a novel \textbf{PMI-inspired progressive token-level modulation}. This mechanism transforms sparse sequence-level rewards into fine-grained learning signals, effectively ``denoising'' the training process by amplifying high-quality reasoning paths while suppressing irrelevant tokens. Extensive experiments on the MER-UniBench benchmark show the superiority and effectiveness of our EmoAgent-R1 in sentiment analysis, basic emotion recognition, and fine-grained emotion understanding.



Our contributions can be summarized as follows:
\begin{itemize}[noitemsep, topsep=0pt]
\item We propose \textbf{EmoAgent-R1}, an agentic MLLM framework that dynamically selects specialized reasoning experts based on the input context, breaking the limitations of simplistic, fixed-prompt approaches.
\item We introduce a \textbf{Cold Start Training strategy} to utilize synthetic answer-conditioned CoT data to stabilize the initial learning of reasoning and routing behaviors. 
\item We propose \textbf{P-GRPO}, a novel RL algorithm that leverages PMI-based modulation to achieve fine-grained credit assignment, enhancing optimization stability and reasoning performance. 
\item Extensive experiments on the MER-UniBench benchmark show the superiority and effectiveness of our EmoAgent-R1 in sentiment analysis, basic emotion recognition, and fine-grained emotion understanding.
\end{itemize}

\section{Related Work} 
\subsection{Multimodal Video Emotion Understanding}

The evolution of multimodal video emotion understanding can be categorized into three distinct paradigms: Traditional Deep Learning, Supervised Fine-Tuning (SFT), and Reinforcement Learning (RL) Alignment.
Initially, traditional deep learning methods utilized CNNs or GNNs to model structural dependencies~\cite{ghosal2019dialoguegcn}, yet they lacked the extensive world knowledge of pre-trained backbones. Subsequently, the field shifted towards the Supervised Fine-Tuning (SFT) paradigm, leveraging the strong priors of Pre-trained Language Models (PLMs) and Multimodal Large Language Models (MLLMs). This category encompasses generative frameworks like UniSA~\cite{li2023unisa}, parameter-efficient adapters such as MSE-Adapter~\cite{yang2025mse} and M2SE~\cite{li2025emoverse}, and instruction-tuned MLLMs like AffectGPT~\cite{lianaffectgpt}, all of which align representations through supervised signals. While effective, SFT methods remain constrained by dataset quality and often rely on shallow fusion strategies that limit fine-grained reasoning. To overcome these bottlenecks, recent efforts have integrated Reinforcement Learning (RL) for preference alignment, as demonstrated by R1-Omni~\cite{zhao2025r1}. However, a critical limitation persists across these advancements: these RL approaches still suffer from uniform credit assignment and monolithic policies. These issues make them fail to distinguish the specific contributions of distinct modalities or temporal segments, thereby diluting reasoning signals and hindering the learning of specialized affective strategies.

\subsection{Reinforcement Learning of LLMs}  
Reinforcement Learning (RL) has emerged as a cornerstone for aligning Large Language Models (LLMs) with human preferences. RLHF~\cite{li2023reinforcement} and PPO~\cite{schulman2017proximal} constitute the dominant paradigm. However, PPO is often constrained by substantial memory overhead and optimization instability, primarily stemming from its reliance on auxiliary value function critics. To address these limitations, Group Relative Policy Optimization (GRPO)~\cite{shao2024deepseekmathpushinglimitsmathematical} has been proposed as a critic-free alternative that estimates baselines via group-level statistics. Its high computational efficiency has proven instrumental in scaling advanced reasoning models such as DeepSeek-R1~\cite{guo2025deepseek}. 
Subsequent research has further advanced GRPO to address complex reasoning tasks. To mitigate the optimization instability inherent in long-horizon trajectories, variants such as DAPO~\cite{yu2025dapo} and BAPO~\cite{xi2025bapo} introduce decoupled or adaptive clipping mechanisms. Furthermore, to counteract length bias, where models exploit reward mechanisms through verbosity, Dr. GRPO~\cite{liu2025understanding} incorporates targeted regularization strategies. Notably, GiGPO~\cite{feng2025group} has begun investigating grouped interventions to enhance credit assignment for specific reasoning steps.

Despite these advancements in text-centric reasoning, a critical gap remains in the domain of multimodal video emotion understanding. Unlike textual logic, which is typically sequential and explicit, emotional cues in video are sparse, asynchronous, and highly localized. It often manifests as fleeting micro-expressions or subtle vocal inflections embedded within predominantly neutral frames. Conventional RL approaches generally rely on sequence-level rewards, broadcasting a uniform learning signal across the entire trajectory. This uniformity reward bias makes an agent fail to differentiate critical affective evidence from irrelevant noise, resulting in inefficient exploration. Consequently, there is a pressing need to develop a progressive, fine-grained credit assignment method specifically tailored to the heterogeneous nature of multimodal data.

\begin{figure*}[htbp]
    \centering
    \includegraphics[width=0.90\linewidth]{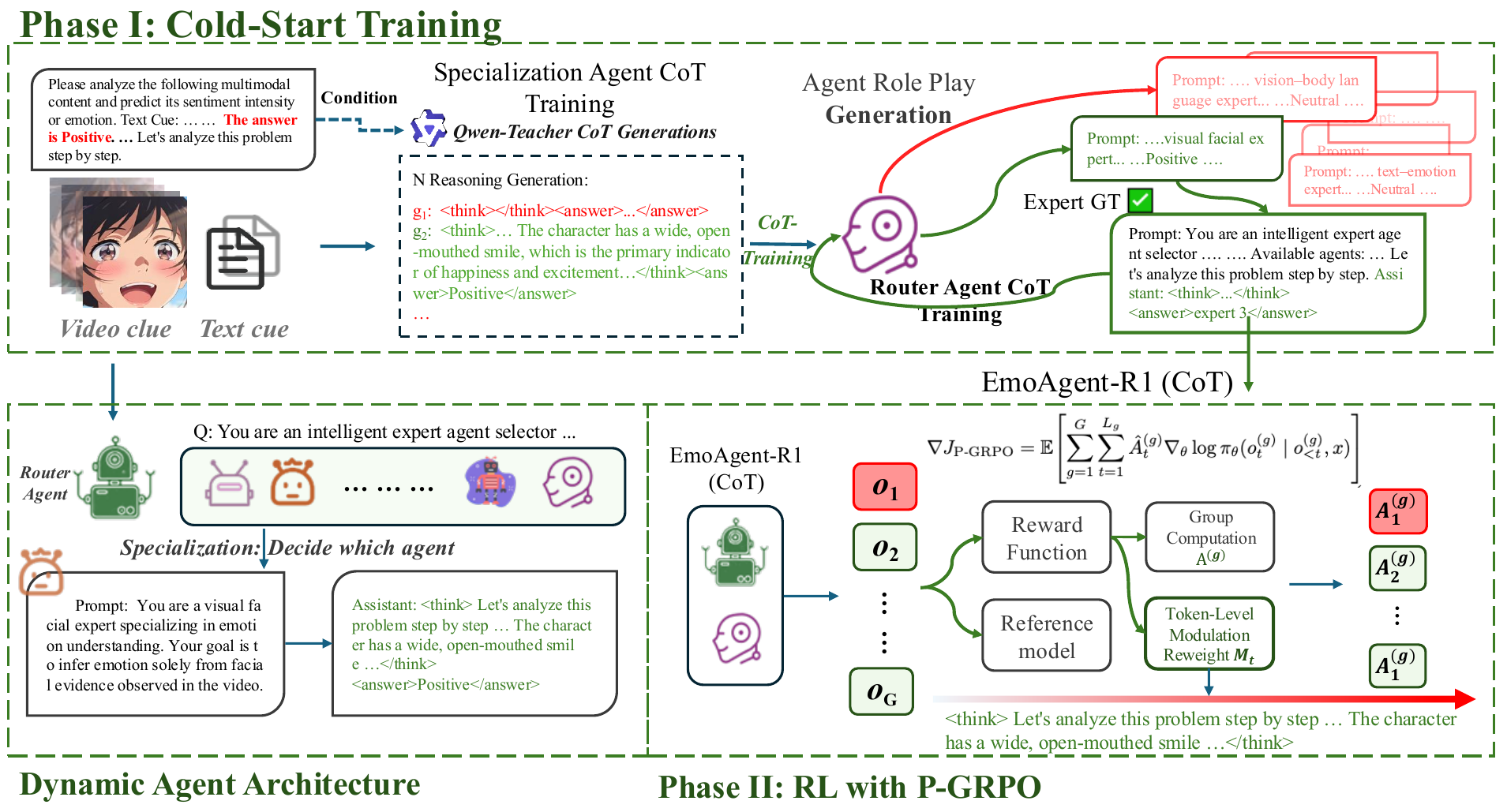}
    \caption{\textbf{The framework of EmoAgent-R1.} The pipeline consists of two phases: (1) \textbf{Cold-Start Training}, where the router and specialized agents are initialized using synthetic answer-conditioned CoT data; and (2) \textbf{RL with P-GRPO}, where the model is further optimized via reinforcement learning with token-level modulation to achieve fine-grained credit assignment.}
    \label{fig:framework}
\end{figure*}
\section{EmoAgent-R1}
\label{sec:method}

We propose \textbf{EmoAgent-R1}, a framework designed to enhance multimodal emotion understanding through dynamic agent specialization and fine-grained reinforcement learning. As illustrated in Figure~\ref{fig:framework}, the EmoAgent-R1 utilizes a unified Multimodal Large Language Model (MLLM) as the backbone of the dynamic agent architecture. The training pipeline is unfolded in two phases: (1) a \textbf{Cold-Start Initialization} phase, which uses synthetic answer-conditioned Chain-of-Thought (CoT) data to establish preliminary emotion reasoning and agent routing capabilities, and (2) a \textbf{Reinforcement Learning} phase, which uses our novel \textbf{Progressive Group Relative Policy Optimization (P-GRPO)} to refine the agent's specialization and emotion reasoning logic under sparse rewards.

\subsection{Problem Formulation}
Given a multimodal input $x = (v, t)$ consisting of video frames $v$ and textual content $t$, our goal is to generate a reasoning trajectory $o = (o_1, \dots, o_L)$ that concludes with the correct emotion prediction $y$. We model this process as a policy $\pi_\theta(o|x)$. Unlike traditional methods that use a static system prompt, we introduce a latent variable $z \in \mathcal{Z}$ representing the choice of a specialized emotion reasoning expert, e.g., a facial expression expert, a sarcasm expert. The emotion reasoning experts are implemented with a different prompt template focusing on different emotional perspectives. The generation process is thus factorized into two steps: expert selection $z \sim \pi_\theta(\cdot|x)$ and expert-guided emotion reasoning $o \sim \pi_\theta(\cdot|x, z)$.

\subsection{Dynamic Agent Specialization Architecture}
\label{sec:architecture}

To dismantle the ``uniformity bias'' inherent in static MLLM prompting, we propose a \textbf{Dynamic Agent Specialization (DSA) Architecture}. The core insight is that complex emotion understanding demands \textit{adaptive reasoning pathways} rather than a monolithic solution. Thus, we implement an agentic workflow that dynamically orchestrates specialized roles within a unified MLLM backbone, transforming the intractable problem of global multimodal understanding into manageable, specialized sub-tasks. 
Central to the DSA framework is the \textbf{Specialized Agent Library}, denoted as $\mathcal{A} = \{a_1, \dots, a_K\}$. Each agent $a_k$ is conceptualized not merely as a prompt template, but also as a distinct \textit{reasoning strategy} designed to activate a specific subset of the model's latent knowledge. By conditioning the generation on specific inductive biases, such as focusing on facial micro-expressions, detecting linguistic sarcasm, or resolving cross-modal contradictions, each agent can isolate critical evidence from noise, thereby enhancing the signal-to-noise ratio for downstream reasoning.

The inference process operates as a \textbf{Two-Stage Agentic Workflow}, formally decoupling strategy selection from execution. Given an input $x$, the MLLM model first functions as a \textit{Routing Agent} $\pi_{\text{router}}(z|x)$. Rather than predicting emotions directly, the routing agent performs meta-cognition by evaluating the input's ambiguity and modality dominance to select the optimal reasoning strategy $z^* \in \mathcal{A}$. Subsequently, the MLLM model is converted to a \textit{Specialized Agent} $\pi_{\text{agent}}(o|x, z^*)$ according to the selected strategy $z^*$. Conditioned on the selected strategy, the agent generates a focused Chain-of-Thought (CoT) trajectory $o$ with as few hallucinations as possible by explicitly narrowing the reasoning scope to the most relevant emotional cues. This hierarchical structure serves as the foundation for our P-GRPO algorithm, enabling precise credit assignment to both routing decisions and reasoning steps.
\subsection{Phase I: Cold-Start Training}  
\label{sec:cold_start}

Reinforcement learning with sparse sequence-level rewards is highly sensitive to initialization. This is particularly critical in our multi-agent framework: if the router explores randomly and agents lack basic reasoning skills, the joint policy may never converge to a high-reward region. To mitigate this issue and establish basic emotion reasoning and agent routing capabilities for backbone MLLM, we construct a cold-start supervised dataset to train both the \textbf{Specialized Agents} and the \textbf{Routing Agent} preliminarily.

\subsubsection{Reasoning Corpus Construction}
We first aggregate a diverse set of multimodal reasoning tasks to endow agents with broad perceptual and cognitive capabilities. The reasoning corpus contains two parts. The first part is \textbf{General Reasoning Data ($N_{gen}=110k$)}, which consists of LLaVA-Video-178k~\cite{zhang2024video}, NeXT-QA~\cite{xiao2021next}, PerceptionTest~\cite{patraucean2023perception}, CLEVRER~\cite{CLEVRER2020ICLR}, and STAR~\cite{wu2021star}. The general reasoning data provide a foundation for temporal reasoning, causal understanding, and cross-modal grounding. The second part is \textbf{Emotion Reasoning Data ($N_{emo}=14k$)}, which contains samples from the training set of MOSEI~(3k)~\cite{zadeh2018multimodal} and Mer-Caption+~(11k)~\cite{lianaffectgpt}. These data are used to specialize the model for affective computing tasks. Unlike generic QA, these samples require attributing emotional states to specific cues, e.g., facial micro-expressions, vocal intonation, aligning directly with our target domain. 

\subsubsection{Answer-Conditioned CoT Emotional Reasoning Synthesis}
To bridge the gap between ground-truth labels and reasoning logic, we employ an answer-conditioned synthesis pipeline. We prompt a teacher Qwen2.5-VL-72B model to generate Chain-of-Thought (CoT) rationales according to the ground-truth emotion. To ensure high-quality synthesis, we carry out a \textbf{DeepSeek-style Rejection Sampling} as following steps: \textbf{1) Generation}: For each sample, the teacher generates $K=8$ candidate trajectories. \textbf{2) Verification}: We employ a verifier based on GPT-4o to filter out trajectories containing factual contradictions or hallucinations, e.g., referencing non-existent visual objects. \textbf{3) Selection}: We only retain those trajectories that logically deduce the correct label and are verified as faithful to the video content.
With these steps, we can yield a high-quality dataset $\mathcal{D}_{\text{reason}}$ for training the Specialized Agents.

\subsubsection{Expert Routing Corpus Synthesis}
Crucially, we must also initialize the Routing Agent $\pi_{\text{router}}(z|x)$ to prevent random expert agent selection during early RL. Since explicit labels for the best expert agent do not exist, we construct an \textbf{Empirical Oracle} via \textit{Hindsight Relabeling}.
For each training instance in $\mathcal{D}_{\text{reason}}$, we first execute all $K$ specialized agents in parallel. Then, we identify the subset of agents $\mathcal{Z}^* \subset \mathcal{A}$ that successfully predict the ground-truth label. Subsequently, we construct routing samples $(x, z)$ where $z \in \mathcal{Z}^*$. This way can effectively transform the unsupervised routing problem into a supervised classification task and teach the router agent to predict: ``\textit{Which expert agent is empirically capable of solving this instance?}''

\subsubsection{Cold-Start Training}
The final cold-start model is obtained by supervised fine-tuning (SFT) on the combined dataset of reasoning trajectories for expert agents and routing decisions for the router agent. With cold-start training, EmoAgent-R1 can establish a robust policy manifold to enable effective exploration in the subsequent reinforcement learning phase.

\subsection{Phase II: Reinforcement Learning}
\label{sec:pgrpo}

\subsubsection{The Formulation of RL with GRPO}
We consider sequence-level reinforcement learning for multimodal reasoning under sparse rewards. Given an input prompt $x$ (e.g., a multimodal video), the policy $\pi_\theta$ generates a trajectory $o = (o_1, \dots, o_L)$ autoregressively. A scalar terminal reward $r(x,o)$ is produced by evaluating the overall correctness, coherence, or emotional consistency of the entire reasoning trajectory $o$. 
To reduce reward variance and enable relative comparison, Group Relative Policy Optimization (GRPO)~\cite{shao2024deepseekmathpushinglimitsmathematical} adopts a group-based sampling strategy. For each prompt $x$, GRPO samples a group of $G$ trajectories $\{o^{(g)}\}_{g=1}^G \sim \pi_\theta(\cdot \mid x)$ and evaluate their terminal rewards jointly. To measure how much better a trajectory is compared to its peers, GRPO deploys a group-relative advantage as follows:
\begin{equation}
A^{(g)} = r^{(g)} - \frac{1}{G}\sum_{j=1}^G r^{(j)},
\end{equation}
The model will be updated according to the following policy gradient estimation:
\begin{equation}
\nabla J_{\mathrm{GRPO}} =
\mathbb{E}\!\left[
\sum_{t=1}^{L_g}
A^{(g)}
\nabla_\theta
\log \pi_\theta(o^{(g)}_t \mid o^{(g)}_{<t}, x)
\right].
\end{equation}

\subsubsection{Progressive Group Relative Policy Optimization}

Despite its variance reduction advantage, GRPO introduces an inherent \emph{uniform credit assignment bias} that the scalar advantage $A^{(g)}$ is broadcast uniformly to all tokens within the trajectory because it implicitly assumes that each token contributes equally to the final outcome. Formally, this is approximated with the token-level action-value function $Q(o_t \mid o_{<t}, x)$ by a constant surrogate $r(x,o)$. It ignores the temporal structure of reasoning and causes those tokens unrelated to reasoning (e.g., syntactic fillers) to receive non-informative gradients while diluting signals for critical reasoning steps.

To address this issue, we propose \textbf{Progressive Group Relative Policy Optimization (P-GRPO)} to decompose the policy gradient into two orthogonal components: (1) an \emph{inter-sample} signal capturing relative quality across completions, and (2) an \emph{intra-sample} signal capturing the progressive contribution of individual tokens. Concretely, we define a step-aware advantage as follows:
\begin{equation}
\hat{A}^{(g)}_t = A^{(g)} \cdot M^{(g)}_t,
\end{equation}
where $M^{(g)}_t$ is a non-negative modulation factor that redistributes the sequence-level advantage across time. The resulting gradient estimator can be defined as follows:
\begin{equation}
\nabla J_{\mathrm{P\text{-}GRPO}} =
\mathbb{E}\!\left[
\sum_{g=1}^G \sum_{t=1}^{L_g}
\hat{A}^{(g)}_t
\nabla_\theta
\log \pi_\theta(o^{(g)}_t \mid o^{(g)}_{<t}, x)
\right].
\end{equation}
When $M^{(g)}_t \equiv 1$, P-GRPO reduces exactly to GRPO.

Ideally, the modulation term $M^{(g)}_t$ should reflect the marginal contribution of token $o_t$, which can be expressed as the incremental pointwise mutual information (PMI) between the token and the success event. Since direct estimation is computationally prohibitive, we construct a low-variance approximation.
Within each group, we normalize rewards as $z^{(g)}=(r^{(g)}-\mu)/(\sigma+\varepsilon)$ and map them to proxy success probabilities $p^{(g)}_{\mathrm{final}}=\sigma(\kappa z^{(g)})$ with $\kappa=2$. Assuming that along a correct reasoning trajectory, confidence in success increases monotonically, prefix success probabilities for a trajectory of length $L_g$ can be defined as follows:
\begin{equation}
p^{(g)}_t =
\mathrm{clip}\!\left(
p^{(g)}_{\mathrm{final}}
\left(\frac{t}{L_g}\right)^{\beta},
\; p_{\min},\; p_{\max}
\right),
\end{equation}
where $\beta=\tfrac{1}{2}$ and $(p_{\min},p_{\max})=(0.1,0.99)$. The incremental log-ratio $\mathrm{PMI}^{(g)}_t = \log (p^{(g)}_t / p^{(g)}_{t-1})$ serves as a tractable surrogate for the ideal incremental PMI.
To ensure non-negative and well-scaled gradients, we convert PMI values into modulation weights $w^{(g)}_t = \exp(\alpha\,\mathrm{clip}(\mathrm{PMI}^{(g)}_t,-c,c))$ with $\alpha=0.5$ and $c=2$. Then, the weights $w^{(g)}_t$ are normalized within each step in a trajectory:
\begin{equation}
M^{(g)}_t =
\frac{w^{(g)}_t}{\frac{1}{L_g}\sum_{i=1}^{L_g} w^{(g)}_i}.
\end{equation}
For unsuccessful trajectories ($p^{(g)}_{\mathrm{final}}\le\tau$),  a constant attenuation factor $M^{(g)}_t=\omega_0$ is applied to suppress noisy gradients.

Finally, we optimize the policy model using a PPO-style clipped objective with KL regularization as follows:
\begin{equation}
\begin{split}
\mathcal{L}(\theta) = \mathbb{E} \Big[ & \min \left(
\rho_{g,t}\hat{A}^{(g)}_t,
\mathrm{clip}(\rho_{g,t},1-\epsilon,1+\epsilon)\hat{A}^{(g)}_t
\right) \Big] \\
& - \beta_{\mathrm{KL}} D_{\mathrm{KL}}(\pi_\theta \| \pi_{\mathrm{ref}}),
\end{split}
\end{equation}
where $\rho_{g,t} =\pi_\theta(o^{(g)}_t)/\pi_{\theta_{\mathrm{old}}}(o^{(g)}_t)$. This objective preserves the stability guarantees of PPO while enabling fine-grained, reward-aligned credit assignment without introducing auxiliary value networks.

\subsubsection{Reward Modeling} 
In our framework, the RLVR training process optimizes the SFT model for emotion recognition using verifiable reward functions. Inspired by DeepSeek R1, we decompose the reward into two components: \textbf{accuracy reward} ($R_{\text{acc}}$) and \textbf{format reward} ($R_{\text{format}}$). The total reward is defined as $\mathcal{R} = \mathcal{R}_{\text{acc}} + \mathcal{R}_{\text{format}}$.
To ensure structured outputs, the \textbf{format reward} is defined as a binary reward function  that evaluates whether the model's response contains both thinking and answer components:
\begin{equation}
\mathcal{R}_{\text{format}}(o|v, q, y) = 
\begin{cases} 
1, & \text{if is\_required\_format}(o) \\
0, & \text{otherwise}
\end{cases}
\end{equation}

The \textbf{accuracy reward} is computed by utilizing the standard OV-MER metrics~\cite{lianov}. Unlike discriminative MER relied on a fixed label set, OV-MER operates within an unconstrained label space. We employ Emotion Wheel (EW)-based metrics that leverage the hierarchical structure of emotion wheels for clustering:
\begin{equation}
\mathcal{R}_{\text{accuracy}}(o|v, q, y) = \text{EW}\left(o^a, y\right).
\end{equation}

\textbf{The concrete design of each specialized agent and the theoretical analysis of our P-GRPO can be found in the Supplementary Materials}.

\begin{table}[t]

\centering
\caption{Dataset statistics for our two-phase training pipeline. We partition samples for Cold-Start (CoT) Training and Reinforcement Learning (RL). The final RL training set was further refined by filtering out noisy instances for which consistent, high-quality reasoning paths could not be generated.}
\label{tab:dataset_stats}
\begin{tabular}{lrrr}
\toprule
\textit{Dataset} & {CoT} & {RL} & {Total} \\
\midrule
General Data & 110k & 15 k & 125 k \\
MER-Caption+ & 11k & 18 k & 29 k \\
MOSEI        & 3k  & 12 k & 15 k \\
\midrule
{Total} & {124 k} & {45 k} & {169 k} \\
\bottomrule
\end{tabular}
\end{table}

\begin{table*}[htbp]
\centering
\definecolor{rankone}{gray}{0.8} 
\definecolor{ranktwo}{gray}{0.92} 
\caption{\textbf{Main results on MER-UniBench.}
MER-UniBench is a unified benchmark covering sentiment analysis, basic emotion recognition, and fine-grained emotion understanding.
All models are evaluated under identical input modalities.
Overall, EmoAgent-R1 consistently outperforms existing baselines and establishes new state-of-the-art results. \colorbox{rankone}{Best} and \colorbox{ranktwo}{second-best} results are highlighted.}
\renewcommand{\arraystretch}{1.2}

\label{tab:mer-unibench}
\resizebox{0.8\textwidth}{!}{
\begin{tabular}{lcccccccccc}
\toprule
\multirow{2}{*}{Model} &
\multicolumn{4}{c}{Sentiment} &
\multicolumn{4}{c}{Basic Emotion} &
Fine-grained & Mean \\
\cmidrule(lr){2-5} \cmidrule(lr){6-9}
& MOSI & MOSEI & SIMS & SIMS v2 & MER23 & MER24 & MELD & IEMO & OV-MERD+ &  \\
\midrule
Otter~\cite{11007678} & 52.89 & 50.44 & 57.56 & 53.12 & 16.41 & 14.65 & 22.57 & 29.08 & 16.63 & 34.82 \\
Video-LLaVA~\cite{lin2024video} & 56.37 & 61.64 & 53.28 & 57.45 & 36.93 & 30.25 & 30.73 & 38.95 & 34.00 & 44.40 \\
PandaGPT~\cite{su-etal-2023-pandagpt} & 58.50 & 64.25 & 62.07 & 65.25 & 39.13 & 47.16 & 38.33 & 47.21 & 35.07 & 50.77 \\
Video-ChatGPT~\cite{Maaz2023VideoChatGPT} & 54.42 & 63.12 & 64.82 & 65.80 & 44.86 & 46.80 & 37.33 & 56.83 & 39.80 & 52.64 \\
VideoChat2~\cite{li2024mvbench} & 66.84 & 54.32 & 69.49 & 70.66 & 33.67 & 54.50 & 36.64 & 48.70 & 39.21 & 52.67 \\
LLaMA-VID~\cite{li2024llamavid} & 61.78 & 63.89 & 69.35 & 67.48 & 50.72 & 57.60 & 42.75 & 46.02 & 45.01 & 56.07 \\
VideoChat~\cite{li2024videochatchatcentricvideounderstanding} & 65.13 & 63.61 & 69.52 & 72.14 & 48.73 & 57.30 & 41.11 & 48.38 & 44.52 & 56.71 \\
Chat-UniVi~\cite{jin2023chatunivi} & 54.53 & 63.18 & 68.15 & 66.36 & 57.62 & 65.67 & 45.61 & 52.37 & 48.00 & 57.94 \\
mPLUG-Owl~\cite{ye2024mplugowlmodularizationempowerslarge} & 72.40 & 72.91 & 72.13 & 75.00 & 56.86 & 59.89 & 49.11 & 55.54 & 48.18 & 62.45 \\
\midrule
AffectGPT~\cite{lianaffectgpt} & \cellcolor{ranktwo}{82.39} & \cellcolor{ranktwo}{81.57} & \cellcolor{ranktwo}{87.20} & \cellcolor{ranktwo}{86.29} & 74.58 & 75.29 & 57.63 & 62.19 & 61.65 & 74.31 \\
{AffectGPT-R1}~\cite{lian2025affectgptr1leveragingreinforcementlearning} & 78.78 & 79.07 & 85.91 & 85.85 & \cellcolor{ranktwo}{77.72} & \cellcolor{ranktwo}{85.29} & \cellcolor{ranktwo}{61.09} & \cellcolor{rankone}{67.42} & \cellcolor{ranktwo}{62.42} & \cellcolor{ranktwo}{75.95} \\
\midrule
{EmoAgent-R1}~(Ours) & \cellcolor{rankone}{83.09~$\uparrow$} & \cellcolor{rankone}{82.66}~$\uparrow$ & \cellcolor{rankone}{87.75}~$\uparrow$ & \cellcolor{rankone}{87.24}~$\uparrow$ & \cellcolor{rankone}{79.08}~$\uparrow$ & \cellcolor{rankone}{86.31}~$\uparrow$ & \cellcolor{rankone}{63.26}~$\uparrow$ & \cellcolor{ranktwo}{66.96}~$\downarrow$ & \cellcolor{rankone}{64.29}~$\uparrow$ & \cellcolor{rankone}{77.85}~$\uparrow$ \\
\bottomrule
\end{tabular}
}
\end{table*}

\begin{table*}[t]

\centering
    \caption{Performance of different models on General benchmarks. \textbf{EmoAgent-R1} consistently outperforms its backbone (Qwen2.5-VL-7B) across reasoning and general understanding tasks, demonstrating strong generalization capabilities.}
\renewcommand{\arraystretch}{1.2}
\resizebox{0.7\textwidth}{!}{%
    \begin{tabular}{lccc| ccc}
    \toprule
    \multicolumn{1}{c}{\multirow{2}{*}{Models}}& \multicolumn{3}{c}{Video Reasoning Benchmark} & \multicolumn{3}{c}{Video General Benchmark} \\
    \cmidrule(lr){2-4} \cmidrule(lr){5-7}
      & VSI-Bench & VideoMMMU & MMVU (mc) & MVBench & TempCompass & VideoMME \\
    \midrule
    GPT-4o~\cite{openai2024gpt4ocard}& 34.0 & 61.2 & 75.4 & - & - & 71.9 \\
    \midrule
    LLaMA-VID~\cite{li2024llamavid}& - & - & - & 41.9 & 45.6 & - \\
    VideoLLaMA2~\cite{cheng2024videollama}& - & - & 44.8 & 54.6 & - & 47.9 \\
    LongVA-7B~\cite{zhang2024longva}& 29.2 & 23.9 & - & - & 56.9 & 52.6 \\
    VILA-1.5-8B~\cite{liu2024nvila}& 28.9 & 20.8 & - & - & 58.8 & - \\
    VILA-1.5-40B~\cite{liu2024nvila}& 31.2 & 34.0 & - & - & - & 60.1 \\
    Video-UTR-7B~\cite{yu2025unhackable}& - & - & - & 58.8 & 59.7 & 52.6 \\
    LLaVA-OneVision-7B~\cite{lillava}& 32.4 & 33.8 & 49.2 & 56.7 & - & 58.2 \\
    Kangaroo-8B~\cite{liu2024kangaroopowerfulvideolanguagemodel}& - & - & - & 61.1 & 62.5 & 56.0 \\
    Qwen2.5-VL-7B~\cite{bai2025qwen2}& 30.1&  48.1& 60.0&  59.0&72.6& 56.6\\
    \midrule
    Video-R1-7B~\cite{feng2025videor}& 34.6 & 49.8 & {64.2} & 62.7 & 72.6 & 57.4 \\
    EmoAgent-R1~(Ours)  & 33.1   & 49.2  &  61.9 &59.4  & 72.6  &  57.1 \\
    \bottomrule
    \end{tabular}%
}
\label{tab:bench}

\end{table*}

\section{Experiments}
\subsection{Experiment Settings}
\noindent\textbf{Datasets.}
Our primary goal is to better accomplish the three typical tasks, including sentiment analysis, basic emotion recognition, and fine-grained emotion detection. Therefore, we incorporate the training set of both the MER-Caption+~\cite{lianaffectgpt} and the MOSEI~\cite{zadeh2018multimodal} datasets to balance the distribution of different answer types across tasks. As detailed in Table~\ref{tab:dataset_stats}, we partition the collected data into 2 distinct subsets to support our two-phase training pipeline: Cold-Start (CoT) and Reinforcement Learning (RL). The CoT subset (124k samples) is utilized to initialize the reasoning and routing capabilities of the agents while maintaining general ability. For the RL phase, we construct another dataset with 45k samples to optimize the policy. Crucially, to ensure the stability of reinforcement learning, we apply a rigorous filtering mechanism to the RL dataset: instances where the teacher model fails to generate consistent, high-quality reasoning paths are discarded as noise. This results in a final high-quality corpus of 169k samples in total.

To evaluate the capabilities of our EmoAgent-R1, we adopt MER-UniBench~\cite{lianaffectgpt} as the benchmark. The MER-UniBench is a unified benchmark with evaluation metrics tailored for typical MER tasks and the free-form, natural language output style of MLLMs. It covers sentiment analysis tasks, basic emotion recognition tasks, and fine-grained emotion understanding tasks.

\noindent\textbf{Metrics.}
We adopt the same metrics in MER-UniBench~\cite{lianaffectgpt} to evaluate model performance. For fine-grained emotion recognition, we adopt the Emotion Wheel (EW)-based metric~\cite{lianaffectgpt} by considering the open-vocabulary nature of the task. This metric maps the predicted emotion $\hat{y}$ and ground truth $y$ into a hierarchical emotional space to calculate semantic similarity. For basic emotion recognition, we adopt the hit rate~\cite{lianaffectgpt} as the evaluation metric, which is set to 1 if the label $y$ is included in the MLLM output and 0 otherwise. For sentiment analysis,  we adopt the weighted average F-score (WAF) as the evaluation metric.


\noindent\textbf{Baselines.}
We compare our EmoAgent-R1 with the following baselines:
Otter~\cite{11007678}, Video-LLaVA~\cite{lin2024video}, PandaGPT~\cite{su-etal-2023-pandagpt}, Video-ChatGPT~\cite{Maaz2023VideoChatGPT}, VideoChat2~\cite{li2024mvbench},
LLaMA-VID~\cite{li2024llamavid}, VideoChat~\cite{li2024videochatchatcentricvideounderstanding}, Chat-UniVi~\cite{jin2023chatunivi}, mPLUG-Owl~\cite{ye2024mplugowlmodularizationempowerslarge},
AffectGPT~\cite{lianaffectgpt}, and {AffectGPT-R1}~\cite{lian2025affectgptr1leveragingreinforcementlearning}. 
\textbf{Please refer to the Supplementary Materials for their brief introduction}.

\noindent\textbf{Implementation Details.}
During training, we set the learning rate to $1\times10^{-5}$ and the maximum number of epochs to 1. The entire implementation is conducted using PyTorch, and the code is executed on H20 GPU with 96 GB of memory. Due to memory constraints, we use a batch size of 8 for cold-start training and a batch size of 1 for reinforcement learning. For the foundation model of EmoAgent-R1, we adopt the Qwen2.5-VL-7B-Instruct~\cite{bai2025qwen25vltechnicalreport}.
\subsection{Performance on MER-UniBench}
The experiment results on MER-UniBench are shown in Table~\ref{tab:mer-unibench}. We can see that EmoAgent-R1 achieves a new  state-of-the-art with a mean score of \textbf{77.85\%}. This performance not only surpasses the previous best specialist method, AffectGPT-R1 (75.95\%), but also demonstrates a substantial lead over general-purpose MLLMs such as Video-LLaVA (44.40\%) and mPLUG-Owl (62.45\%), validating the efficacy of our reinforcement learning-based dynamic agent specialization method in bridging the gap between general video understanding and affective computing.
In specific tasks, EmoAgent-R1 also exhibits exceptional robustness. For sentiment analysis, EmoAgent-R1 consistently secures the top rank across all four datasets, achieving notable scores of \textbf{83.09\%} on MOSI and \textbf{87.75\%} on SIMS. This indicates a refined ability to integrate heterogeneous signals for precise polarity detection. For basic emotion recognition, EmoAgent-R1 achieves first-class performance on  MER23 (\textbf{79.08\%}), MER24 (\textbf{86.31\%}), IEMO (\textbf{66.96\%}), and MELD (\textbf{63.26\%}), showing that it can effectively distinguish subtle emotional shifts challenging for monolithic models and proving its versatility across diverse emotion taxonomies. On the most challenging task of fine-grained emotion understanding (OV-MERD+), which demands high-level reasoning for open-vocabulary descriptions, EmoAgent-R1 (\textbf{64.29\%}) outperforms AffectGPT-R1 (62.42\%) by over 1.87\%, underscoring EmoAgent-R1's capacity to handle complex emotional semantics and mitigate the uniformity bias, which is a benefit of the fine-grained credit assignment of our P-GRPO. 
Overall, these results fully demonstrate the effectiveness of EmoAgent-R1. 

\section{Generalization on General benchmarks}
As shown in Table~\ref{tab:bench}, \textbf{EmoAgent-R1} maintains robust generalization capabilities on general video benchmarks. Compared to its backbone \textbf{Qwen2.5-VL-7B}, our model achieves steady improvements across most metrics, particularly in reasoning tasks such as VSI-Bench (33.1 vs. 30.1) and MMVU (61.9 vs. 60.0), while preserving performance on general understanding tasks like TempCompass. Furthermore, EmoAgent-R1 performs on par with the reasoning-specialized \textbf{Video-R1-7B}, achieving comparable results across the board and slightly surpassing it on VSI-Bench. These findings indicate that our emotion-specific optimization enhances targeted reasoning skills without compromising the model's fundamental video understanding abilities.

\begin{table}[t]
  \centering
  \caption{\textbf{Ablation Study.} Comparison of SFT, RL, and EmoAgent-R1. Note that P-GRPO is crucial for performance. sCOT denotes Specialization COT,  rCOT denotes Router COT, and DAS denotes dynamic agent specialization.}
  \renewcommand{\arraystretch}{1.2}
  \label{tab:ablation_study_optimized}
  
  \definecolor{rankone}{gray}{0.8} 
  \definecolor{ranktwo}{gray}{0.92} 

  \resizebox{1\linewidth}{!}{%
  \begin{tabular}{llcccc}
    \toprule
    
    \multirow{3}{*}{\textit{Stage}} & \multirow{3}{*}{\textit{Module}}& \multicolumn{2}{c}{\textbf{Sentiment}} & \textbf{Basic} & \textbf{Fine-} \\
    && \multicolumn{2}{c}{\textbf{Analysis}} & \textbf{Emo.} & \textbf{grained} \\
    
    \cmidrule(lr){3-4} \cmidrule(lr){5-5} \cmidrule(lr){6-6}
     && {MOSEI} & {SIMS v2} & {MER23} & {OV+} \\
    \midrule
    \multirow{3}{*}{\textit{CoT}}&\textit{Baseline} & 66.09 & 74.47& 57.18 & 49.06 \\
    &\textit{+ sCOT} & 72.50 & 79.21 &66.42 & 56.02 \\
    &\textit{+ rCOT~(w / DAS)} & 79.07 & 82.30& 74.70& 61.09 \\
    \midrule
    \multirow{4}{*}{\textit{RL}}
    &\textit{w / GRPO}   & 81.11& 85.30 & 77.13& 63.35\\
    &\textit{w / P-GRPO} & \cellcolor{rankone}{82.66} & \cellcolor{rankone}{87.24} & \cellcolor{rankone}{79.08} & \cellcolor{rankone}{64.29} \\
    \cmidrule(lr){2-6}
    &\textit{w / o DAS} & 78.77& 82.11& 75.18& 60.53  \\
    &\textit{w / DAS} & \cellcolor{rankone}{82.66} & \cellcolor{rankone}{87.24} & \cellcolor{rankone}{79.08} & \cellcolor{rankone}{64.29} \\
    \bottomrule
  \end{tabular}%
  }
\end{table}

  
  
    
    

\subsection{Ablation Study}
\label{sec:ablation}

To validate the effectiveness of the core components in EmoAgent-R1, we conduct a comprehensive ablation study focusing on the two training phases: Cold-Start Training and Reinforcement Learning (RL). The quantitative results are summarized in Table~\ref{tab:ablation_study_optimized}.

\subsubsection{Effectiveness of Cold-Start Strategies}
As demonstrated in the first stage of Table~\ref{tab:ablation_study_optimized}, the vanilla baseline model exhibits limited performance on complex multimodal tasks (66.09\% on MOSEI). The introduction of \textit{Specialization CoT} significantly improves the model's reasoning capabilities, raising the MOSEI score to 72.50\%. Crucially, the further integration of \textit{Router CoT} (w/ Dynamic Agent Specialization) provides a substantial performance improvement across all metrics, e.g., 6.57\% gain on MOSEI and 5.07\% gain on OV+ compared to Specialization CoT. This indicates that initializing the router's meta-cognitive ability to select appropriate experts is a prerequisite for effective downstream optimization.

\subsubsection{The Superiority of P-GRPO}
In the RL stage, we compare our proposed Progressive Group Relative Policy Optimization (P-GRPO) against the standard GRPO baseline. Although standard GRPO yields improvements over the SFT model by optimizing for sequence-level rewards, P-GRPO consistently outperforms it, achieving state-of-the-art results (82.66\% on MOSEI and 64.29\% on OV+). This validates that our PMI-inspired token-level modulation effectively mitigates the coarse-grained credit assignment issue, enabling the model to distinguish critical reasoning steps from irrelevant noise during optimization.

\subsubsection{Necessity of Dynamic Agent Specialization Architecture}
Finally, we analyze the impact of the dynamic agentic workflow. By removing the dynamic agent mechanism and reverting to a monolithic policy (``w/o Dynamic Agent Specialization''), we observe a sharp decline in performance (e.g., OV+ drops from 64.29\% to 60.53\%). This finding supports our core hypothesis that static, monolithic prompts suffer from ``uniformity bias'' and lack the flexibility required to capture the heterogeneous and dynamic nature of emotional cues in video.

\subsection{Scaling Analysis}
We also investigate the impact of model scaling on multimodal emotion understanding by comparing EmoAgent-R1 with different backbone sizes against their corresponding instruction-tuned baselines. The results are shown in Table~\ref{tab:model_scaling}.
Across both Qwen2.5-VL and Qwen3-VL families, we observe that simple parameter scaling yields only marginal gains. For example, increasing the backbone from Qwen2.5-VL-3B to 7B improves the mean score by merely $+0.35$ points (61.35 $\rightarrow$ 61.70). In contrast, with our proposed reinforcement learning-based dynamic agent specialization, parameter scaling can lead to substantial and consistent improvements across all tasks.
Specifically, EmoAgent-R1 (3B) outperforms the Qwen2.5-VL-7B-instruct baseline by a large margin on all benchmarks, achieving a $13.52$ gain on MOSEI, $8.80$ on SIMS v2, $18.73$ gain on MER23, and $11.44$ gain on OV-MERD+. This result highlights that structured reasoning optimization and adaptive agent routing
are significantly more effective than increasing model capacity alone.

Furthermore, scaling EmoAgent-R1 from 3B to 7B continues to provide additional benefits, indicating that our framework is complementary to model scaling rather than a replacement. A similar trend is observed on the Qwen3-VL backbone, where EmoAgent-R1 (4B) achieves strong performance despite using fewer parameters.
Overall, these results demonstrate that dynamic agent specialization reshapes the scaling behavior of MLLMs, enabling smaller models to rival or surpass much larger counterparts through more effective reasoning and credit assignment.

\textbf{Due to space limitations, the qualitative analysis about Overcoming Modality Conflict via Specialization and the analysis about the dynamics of specialized agent selection can be found in the Supplementary Materials.}

    
    

\begin{table}[t]
  \centering
    \caption{\textbf{Scaling Impact.}Scaling behavior of EmoAgent-R1 across different backbone sizes. Smaller EmoAgent-R1 models achieve competitive performance against larger instruct baselines.}
  \label{tab:model_scaling}
    \renewcommand{\arraystretch}{1.2}
  \definecolor{rankone}{gray}{0.8} 
  \definecolor{ranktwo}{gray}{0.92} 
  \resizebox{\linewidth}{!}{%
  \begin{tabular}{lccccc}
    \toprule
    \multirow{2}{*}{\textit{Base Models}} & \multicolumn{2}{c}{{Sentiment}} & {Basic} & {Fine-grained} & \multirow{2}{*}{Mean}\\
    \cmidrule(lr){2-3} \cmidrule(lr){4-4} \cmidrule(lr){5-5}
    & {MOSEI} & {SIMS v2} & {MER23} & {OV-MERD+} \\
    \midrule
    \multicolumn{5}{c}{\textit{{Qwen2.5-VL}}} \\\midrule
    Qwen2.5-VL-3B-Instruct & 65.78 & 74.18 & 56.93 & 48.50 & 61.35\\
    {EmoAgent-R1 (3B)} & \cellcolor{rankone}{79.78} & \cellcolor{rankone}{83.27} & \cellcolor{rankone}{75.91} & \cellcolor{rankone}{60.50} & \cellcolor{rankone}{74.87}\\
    
    Qwen2.5-VL-7B-Instruct & 66.09 & 74.47& 57.18 & 49.06 & 61.70\\
    {EmoAgent-R1 (7B)} & \cellcolor{rankone}{82.66} & \cellcolor{rankone}{87.24} & \cellcolor{rankone}{79.08} & \cellcolor{rankone}{64.29} & \cellcolor{rankone}{78.32}\\
    \midrule
    \multicolumn{5}{c}{\textit{{Qwen3-VL}}} \\\midrule
    Qwen3-VL-4B-Instruct & 66.30& 74.47 & 57.18 & 48.68 & 61.66\\
    {EmoAgent-R1 (4B)} & \cellcolor{rankone}{80.72} & \cellcolor{rankone}{85.20} & \cellcolor{rankone}{77.86} & \cellcolor{rankone}{62.78} & \cellcolor{rankone}{76.64}\\
    \bottomrule
  \end{tabular}%
  }
\end{table}

\section{Conclusion}
\label{sec:conclusion}

We proposed EmoAgent-R1, a two-step agentic workflow framework that redefines multimodal emotion understanding by shifting the paradigm from static, monolithic modeling to Reinforcement Learning-based Dynamic Agent Specialization. 
Our core contribution lies in the agentic workflow that empowers the MLLM to act as an adaptive router to dynamically select a specialized reasoning expert agent based on the unique ambiguity and modality dominance of each input. 
This mechanism effectively breaks the "uniformity bias" inherent in traditional monolithic prompting approaches, allowing the model to align its reasoning strategy with the heterogeneous and fleeting nature of emotional cues. 
To facilitate the learning of such complex behaviors under sparse rewards, we propose P-GRPO to stabilize the optimization of these specialized agents by providing fine-grained credit assignment. 
Extensive experiments on MER-UniBench confirm that this dynamic agent specialization strategy can enhance emotion understanding and reasoning ability, and EmoAgent-R1 establishes new state-of-the-art performance, paving the way for more adaptive, agent-centric affective computing systems.

\nocite{langley00}
\bibliographystyle{ACM-Reference-Format}
\bibliography{references}


\end{document}